\documentclass[10pt,twocolumn,letterpaper]{article}

\usepackage{iccv}
\makeatletter
\@namedef{ver@everyshi.sty}{}
\makeatother
\usepackage{tikz}

\usepackage{times}
\usepackage{epsfig}
\usepackage{graphicx}
\usepackage{amsmath}
\usepackage{amssymb}
\usepackage{tikz} 
\usepackage{xcolor}
\usepackage{pgfplots}
\pgfplotsset{compat=1.5}
\usepackage{caption}
\usepackage{subcaption}

\usepackage[pagebackref=true,breaklinks=true,letterpaper=true,colorlinks,bookmarks=false]{hyperref}

\usepackage{times}
\usepackage{epsfig}
\usepackage{graphicx}
\usepackage{amsmath,amssymb}
\usepackage{bm,xspace}
\usepackage{verbatim}
\usepackage{multirow}
\usepackage{wrapfig}
\usepackage{balance}
\usepackage{multicol}        
\usepackage{mathrsfs}
\usepackage{mathtools}
\usepackage{bbm}
\usepackage{times}
\usepackage{enumerate}
\usepackage{algorithm} 
\usepackage{tabularx}
\usepackage{array}
\usepackage{bm}
\usepackage{cleveref}
\usepackage[bottom]{footmisc}

\newcommand{\SO}[1]{\ensuremath{\operatorname{SO}(#1)}}
\def \sphericalcnns{Spherical CNNs}
\newcommand{\reconstructed}{\mathcal{S}^\star}
\newcommand{\source}{\mathcal{S}}

\newcommand{\loss}{\mathcal{L}}
\newcommand{\parameters}{\theta}

\def\dd{\mathbf{d}}

\def\pp{\mathbf{p}}

\DeclareMathOperator*{\argmax}{arg\,max}

\definecolor{ours}{rgb}{0.0, 0.0, 0.55}
\definecolor{ours_flare}{rgb}{0.82, 0.1, 0.26}
\definecolor{shot}{rgb}{0.0, 0.75, 1.0}
\definecolor{fpfh}{rgb}{0.0, 0.5, 0.0}
\definecolor{usc}{rgb}{0.0, 0.81, 0.82}
\definecolor{spinimages}{rgb}{0.0, 0.0, 0.0}
\definecolor{ppfnet}{rgb}{0.4, 0.22, 0.33}
\definecolor{ppffoldnet}{rgb}{1.0, 0.6, 0.6}
\definecolor{cgf}{rgb}{1.0, 0.6, 0.6}
\definecolor{3dmatch}{rgb}{1.0, 0.84, 0.0}
\definecolor{atlasnet}{rgb}{1.0, 0.66, 0.07}

\usepackage[pagebackref=true,breaklinks=true,letterpaper=true,colorlinks,bookmarks=false]{hyperref}

\iccvfinalcopy

\ificcvfinal\fi
\begin{document}

\title{Learning an Effective Equivariant 3D Descriptor Without Supervision}

\author{Riccardo Spezialetti, Samuele Salti, Luigi di Stefano\\
	Department of Computer Science and Engineering (DISI)\\
	University of Bologna, Italy\\
	{\small \{riccardo.spezialetti, samuele.salti, luigi.distefano \}@unibo.it}
}
\maketitle

\begin{abstract}
Establishing correspondences between 3D shapes is a fundamental task in 3D Computer Vision, typically addressed by matching local descriptors. Recently, a few attempts at applying the deep learning paradigm to the task have shown promising results. Yet, the only explored way to learn rotation invariant descriptors has been to feed neural networks with highly engineered and invariant representations provided by existing hand-crafted descriptors, a path that goes in the opposite direction of end-to-end learning from raw data so successfully deployed for 2D images. 

In this paper, we explore the benefits of taking a step back in the direction of end-to-end learning of 3D descriptors by disentangling the creation of a robust and distinctive rotation equivariant representation, which can be learned from unoriented input data, and the definition of a good canonical orientation, required only at test time to obtain an invariant descriptor. To this end, we leverage two recent innovations: spherical convolutional neural networks to learn an equivariant descriptor and plane folding decoders to learn without supervision.
The effectiveness of the proposed approach is experimentally validated by outperforming hand-crafted and learned descriptors on a standard benchmark. 
\end{abstract}

\section{Introduction}
\label{sec:introduction}
Surface matching is a challenging problem in 3D Computer Vision. It has a large number of applications such as 3D Object Recognition, 3D Object Retrieval, 3D Registration and Reconstruction. 
The definition of compact and effective representations of the local geometry of a surface, usually referred to as \textit{descriptors}, plays a key role in surface matching. Indeed, performance of the algorithms proposed to tackle the above mentioned applications is often largely determined by the effectiveness of the chosen descriptor. This has fostered intensive research in the area of local 3D descriptors in the last decades \cite{tombari2010unique, salti2014shot, johnson1999using, guo2013rotational,rusu2009fast}.

The success of deep neural networks in image recognition has motivated a recent paradigm shift from handcrafted algorithms to data-driven approaches also in the design of local 3D descriptors  \cite{zeng20173dmatch, khoury2017learning, deng2018ppffoldnet, deng2018ppfnet}. 
However, state-of-the art proposals do not actually learn new local 3D descriptors from the input data but from existing handcrafted 3D descriptors, which are already rotation-invariant by design: e.g., CGF \cite{khoury2017learning} starts from a high-dimensional input parameterization which closely resembles the Unique Shape Context (USC) descriptor \cite{tombari2010unique}, while PPF-FoldNet \cite{deng2018ppffoldnet} relies on the well-known Point Pair Features (PPF) \cite{drost2010model}.  In other words, due to the difficulty of feeding neural networks with unorganized input data \cite{khoury2017learning}, these approaches create new descriptors by actually learning how to robustly \textit{compress} a specific invariant handcrafted descriptor. 

We argue that the drawbacks of relying on invariant handcrafted descriptors as input data to feed neural networks are twofold. On one hand, there not exist an optimal handcrafted descriptor across applications and datasets, as vouched by recent evaluations \cite{guo2016comprehensive}.
Therefore, for instance, performance of PPF-FoldNet are limited on some scenarios and datasets by the handcrafted design decision of using PPF as input representation.
On the other hand, to achieve rotation invariance, existing handcrafted descriptors used in deep learning pipelines rely either on the normal at the point as a reference axis \cite{deng2018ppffoldnet} or on a local reference frame (LRF) \cite{khoury2017learning} to express point coordinates and angles with respect to a canonically oriented reference frame. Repeatability of the axis or the LRF directly affects the invariance and robustness of the input descriptor \cite{petrelli2012repeatable, petrelli2011repeatability} and, in turn, of the descriptor learned from such representations.
However, parameters used to obtain such canonical orientation (e.g. the number of neighbours to estimate the normals, how to establish a reference direction on the tangent plane in a LRF, \etc ) are again handcrafted design decisions and are not optimized during training. 

Reliance on rotation-invariant handcrafted descriptors as input representations deviates significantly from the end-to-end learning paradigm so successfully applied to images. Therefore, in this paper we investigate on whether leaving the model free to learn an optimal descriptor from a non-canonically oriented input representation may unleash the untapped potential of deep learning also in this scenario.
To this end, we exploit the paradigm recently proposed  in FoldingNet \cite{yang2018foldingnet}  and AtlasNet \cite{groueix2018papier} to realize unsupervised learning of an embedding space from 3D data, which learns to deform, according to the latent representation, points sampled from a plane so as to reconstruct the input surface. This concept has already been deployed  to obtain  an invariant 3D descriptor by reconstructing the Point Pair Features of the input data \cite{deng2018ppffoldnet}. In our proposal, however, the learned latent space has to encode pose information in order to be able to reconstruct the input under arbitrary poses, as it will be shown later (Sec. \ref{subsec:rotation_equivariant}). We argue that the ability to learn an embedding \textit{equivariant} with respect to rotations of the input is the most sound approach to include pose information in the latent space. To this end, we leverage recent work on Spherical CNNs \cite{cohen2018spherical, esteves2018learning}, which have enhanced the deep learning machinery by enabling it to learn also rotation-equivariant representations from 3D spherical signals by means of correlations defined for the $\SO3$ group of rotations.
Hence, in our architecture, a Spherical CNN encoder learns to summarize the geometry around a feature point into a rotation-equivariant embedding and a decoder warps a 2D grid in order to reconstruct the raw input data. This enables learning of an equivariant embedding without using noisy and arbitrary canonical orientations at training time.

To perform pose invariant descriptor matching at test time, we have investigated two alternative ways to orient our equivariant descriptor: we can again exploit the peculiar nature of the Spherical CNN output, which is a signal living in $\SO3$, to define a canonical orientation directly from the computed embedding; or we can orient the descriptor according to a canonical orientation provided by an external local reference frame computed on the input data. While the first approach enables end-to-end learning of the descriptor and the LRF, we have so far obtained better results with the second one.
In particular, we have validated our claim on the superiority of learning a local descriptor from raw unoriented input data by comparing the two variants against handcrafted and learned methods on the popular 3DMatch benchmark data set \cite{zeng20173dmatch}. Our proposal improves the state-of-the art by a remarkable margin, outperforming the method based on the same unsupervised learning framework, but applied to an invariant descriptor, by more than 0.23 points of fragments registation recall (31\% increase).

\section{Related Work}
\label{sec:related}
This section provides a review of the main proposals in the field of local descriptors, starting from  early hand-crafted methods up to novel approaches based on deep learning.  

\textbf{Hand-crafted 3D Local Descriptors} 
A local 3D descriptor creates a compact representation of a 3D surface by collecting geometric or topological measurements into histograms.  Approaches such as 
\textit{Spin Images} \cite{johnson1999using}, \textit{Unique Shape Context} \cite{tombari2010unique} and \textit{RoPs} \cite{guo2013rotational} rely on the spatial distribution of the points on the surface, while others like \textit{FPFH} \cite{rusu2009fast} and \textit{SHOT} \cite{salti2014shot} exploit geometric properties of the surface such as normals or curvatures. Rotation invariance is achieved using either a Local Reference Frame or a Reference Axis.

\textbf{Learned 3D Local Descriptors}
The impressive progress in image recognition yielded by deep learning has inspired similar approaches to learn descriptor from 3D data. However, the unorganized nature of point clouds makes this extension not straightforward. As a consequence, several parallel tracks regarding the representation of the input data have emerged. Early works represent a 3D object as a collection of 2D views \cite{su2015multi,wei2016dense}. Another approach concerns  dense 3D voxel grids, with voxels containing either a binary occupancy grid \cite{maturana2015voxnet, wu2017sampling} or an alternative representation of the surface \cite{zeng20173dmatch}. To limit the memory occupancy of voxel grids, researchers either rely on coarse spatial resolutions, which, however, introduce artifacts and hinder the ability to learn fine geometric structures, or on space partition methods like k-d trees or octrees \cite{klokov2017escape, tatarchenko2017octree}. Other methods, differently, deploy high-dimensional hand-crafted features to parameterize the input point cloud and then use deep learning to project it into lower dimensional spaces \cite{khoury2017learning,deng2018ppffoldnet}.

\textbf{Learning from Raw 3D Data}   PointNet \cite{qi2017pointnet} and PointNet++ \cite{qi2017pointnet++} are pioneering works presenting a general framework to learn features directly from raw point clouds data. Although yielding excellent performance in point cloud segmentation and classification tasks, these architectures have  not been used yet to perform local surface description, likely due to the inability of providing rotation invariance.  Nonetheless, PointNet is the core building block of PPFNet \cite{deng2018ppfnet}, which relies on  raw point coordinates, normals and Point-Pair Features in order to learn a local feature descriptor. Indeed, due to the reliance on the PointNet architecture, PPFNet is not rotation invariant.

\section{Proposed Method}
\label{sec:method}
\begin{figure*}
	\centering
	\includegraphics[width=0.90\textwidth]{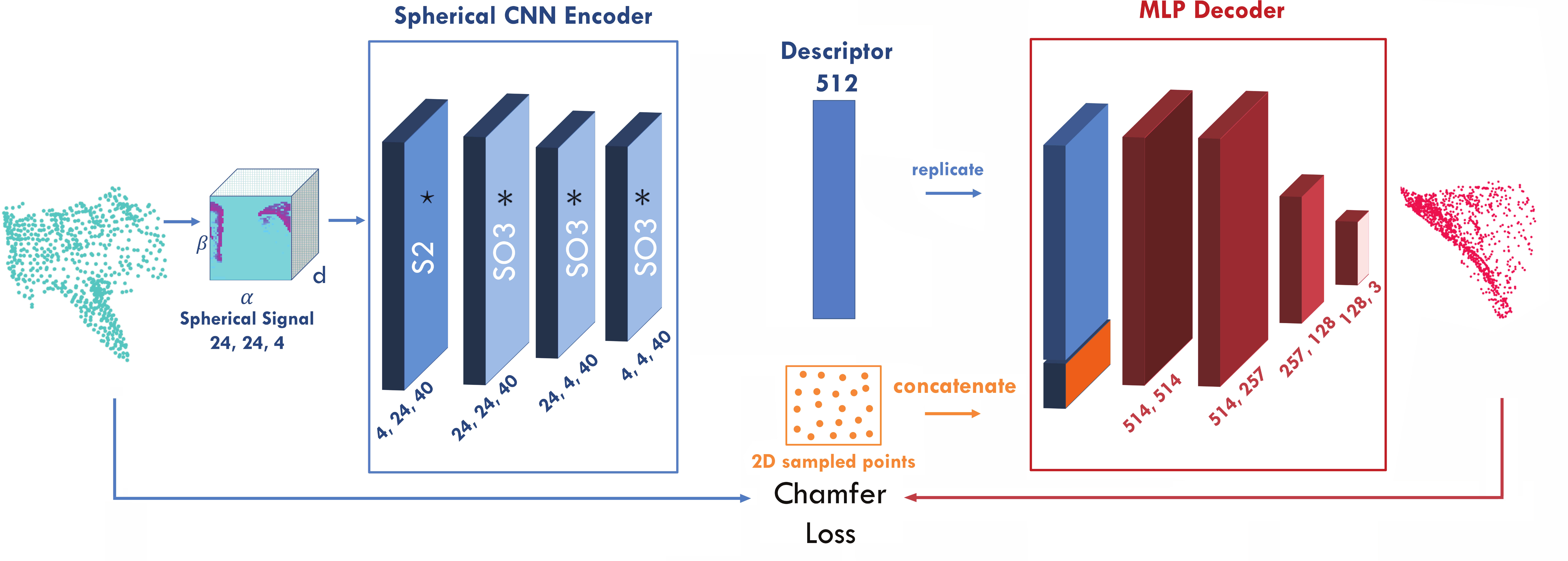}
	\caption{Architecture of the proposed method. The points within the local support of a given feature point $\pp$ are converted into a spherical signal representation, and then sent through the spherical encoder to get an equivariant descriptor. The numbers below the spherical signal indicate the number of cells along $\alpha$, $\beta$ and $d$. The decoder reconstructs the original point cloud deforming sampled 2D points according to the descriptor. Operations in the encoder are implemented through the Generalized Fourier Transform with signals discretized according to a bandwidth parameter \cite{cohen2018spherical}. The triplets below the encoder layers indicate input bandwidth, output bandwidth and number of channels. As for the decoder, the pairs indicate the number of input and output channels, respectively.}
\label{fig:architecture}
\end{figure*}

In this section we present the whole pipeline of our method, graphically illustrated in Figure \ref{fig:architecture}. Please note that our encoder only contains correlation layers, \ie it does not include a max pooling layer at the end to learn a pose-invariant descriptor, which is instead present in the architectures proposed in \cite{cohen2018spherical}.

\subsection{Background}
\label{subsec:background}

As we rely on \sphericalcnns{}, to make the paper self-contained we provide a brief overview of the mathematical model behind it. For more details, please refer to \cite{cohen2018spherical}.

The basic intuition behind Spherical CNNs can be grasped by analogy with the classical planar correlation used by traditional CNNs. As explained in \cite{cohen2018spherical}, the value of the output feature map at $x \in \mathbb{Z}^2$ in a planar correlation can be understood as the inner product between the input feature map and the learned filter shifted by $x$. By analogy, the value of the output feature map at $R \in \SO{3}$ in a spherical correlation can be understood as the inner product between the input feature map and the learned filter, \textit{rotated} by $R$.

A source of confusion when switching from traditional to spherical CNNs is that the space where input signals, for instance point clouds, and feature maps live is different: the former live in $\mathbb{R}^3$, while the latter live in $\SO{3}$. Therefore, when we read the value of a feature map, we are getting the response of the filter for a specific rotation, not for a location in the input cloud. This is not the case with traditional correlations, where both the input images and the feature maps live in $\mathbb{Z}^2$, and the concept of receptive field of a feature map is more intuitive. 

Some useful definitions to understand spherical CNNs also from a formal point of view are given below.

\noindent\textbf{The Unit Sphere} $S^2$ can be defined as the set of points $x \in \mathbb{R}^3$ with norm 1. It is a two-dimensional manifold, which can be parameterized by spherical coordinates $\alpha \in [0, 2\pi]$ (azimuth) and $\beta \in [0, \pi]$ (inclination).

\noindent\textbf{Spherical Signals} the kernels of our Spherical encoder are designed as continuous $K$-valued functions: $f: S^2 \rightarrow \mathbb{R}^K$, where $K$ is the number of channels.

\noindent\textbf{Rotations} A rotation in three dimensions lives in a three-dimensional manifold called $\SO3$, the ``special orthogonal group``. 
As in \cite{cohen2018spherical} the rotation group $\SO3$ can be parameterized by ZYZ-Euler angles $\alpha \in [0, 2\pi], \beta \in [0, \pi], $ and $\gamma \in [0, 2\pi]$. Rotations can be represented by $3\times 3$ matrices that preserve distance (i.e. $\|Rx\| = \|x\|$) and orientation ($det(R) = +1$). If we represent points on the sphere as 3D unit vectors $x$, a rotation can be performed by using the matrix-vector product $Rx$.

\noindent\textbf{Rotations of Spherical Signals} The spherical correlation operator needs to rotate the filters on the sphere. For this purpose, \cite{cohen2018spherical} introduces the operator $L_R$ that takes a function $f$ and produces a rotated function $L_Rf$ by composing $f$ with the rotation $R^{-1}$:

\begin{equation} 
\label{eqn:rot_sphere}
[L_R f](x) = f(R^{-1} x)
\end{equation}

\noindent\textbf{Spherical Correlation} 
Denoting with $\langle \psi, f \rangle$ the inner product on the vector space of spherical signals defined as in \cite{cohen2018spherical}, the correlation between
a $K$-valued spherical signal $f$ and a filter $\psi$, $f, \psi:S^2 \rightarrow \mathbb{R}^K$ can be formalized as:
\begin{equation}
\label{eqn:sphere_conv}
[\psi \star f](R) = \langle L_R \psi, f \rangle = \int_{S^2} \sum_{k=1}^K \psi_k(R^{-1} x) f_k(x) dx.
\end{equation}

This is the operation performed by the first layer of our encoder (Figure \ref{fig:architecture}). Unlike the standard definition of spherical convolution \cite{driscoll1994computing}, which gives as output a function on the sphere $S^2$, the spherical correlation yield a signal on $\SO3$. The use of a conventional convolution definition would limit the expressive capacity of the network due to the symmetry along the Z axis of the learned filters.

\noindent\textbf{Rotation of $\SO3$ Signals}
Similarly to what has been defined for spherical correlation in Eq. \eqref{eqn:sphere_conv}, to define a correlation in $\SO{3}$ the operator in Eq. \eqref{eqn:rot_sphere} must be generalized so that it can act on $\SO3$. For a signal $h : \SO3 \rightarrow \mathbb{R}^K$, and $R, Q \in \SO3$:

\begin{equation}
[L_R h](Q) = h(R^{-1} Q).
\label{eqn:L_R_SO3}
\end{equation}
The term $R^{-1} Q$ in Eq. \eqref{eqn:L_R_SO3} denotes the composition of rotations.

\noindent\textbf{Rotation Group Correlation}
Likewise in Eq. \eqref{eqn:sphere_conv}, we can define the correlation between a signal and a filter on the rotation group, $h, \psi : \SO3 \rightarrow \mathbb{R}^K$, as follows:
\begin{equation}
\label{eqn:SO3_conv}
[\psi \ast h](R) = \langle L_R \psi, f \rangle = \int_{\SO3} \sum_{k=1}^K \psi_k(R^{-1} Q) h_k(Q) dQ.
\end{equation}
This is the operation performed by the all the layers of our encoder but the first one (Figure \ref{fig:architecture}). The integration measure $dQ$ is the invariant measure on $\SO3$, which may be expressed in ZYZ-Euler angles as $d\alpha \sin(\beta) d\beta d\gamma / (8 \pi^2)$.
Please note that unlike in \cite{cohen2018spherical} for better clarity we denote as $\star$ the spherical correlation \eqref{eqn:sphere_conv} while with $\ast$ the rotation group correlation \eqref{eqn:SO3_conv}.

\subsection{Learning from Spherical Signals}
\label{subsec:learning_from_spherical_signal}
Our feature encoder operates on signals defined in a spherical domain. Hence, the local geometry surrounding a feature point needs to be converted into a spherical representation. A common strategy adopted by \cite{cohen2018spherical, esteves2018learning} is to project a 3D mesh onto an enclosing discretized sphere using a  raycasting scheme. Since our input data is not a regular watertight mesh, but a point cloud corresponding to the neighborhood of the point we wish to describe, we first convert 3D points into a spherical coordinate system and then construct a quantization grid in this new coordinate system, similarly to \cite{you2018prin}. The $i$-th cell in the quantization is identified with three spherical coordinates $(\alpha[i], \beta[i], d[i]) \in S^2 \times D $ where $\alpha[i]$ and $\beta[i]$ represent the azimuth and inclination angles of its center and $d[i]$ is the distance from the sphere center. The $K$-valued spherical signal $f: S^2 \rightarrow \mathbb{R}^K$ is then composed by $K$ concentric spheres corresponding to the number of subdivisions along the distance axis, each sphere encoding the density of the points within each cell $(\alpha[i], \beta[i])$ at a given distance $d[k]$. To take into account the non-uniform spacing in the spherical space, cells near the south or north pole are wider in spherical coordinates, as discussed in \cite{you2018prin}.

A spherical signal is computed on the local neighborhood of every input point we wish to describe (i.e., every \textit{keypoint}). The signal then goes through our architecture to learn an equivariant bottleneck layer, which can then be used as a descriptor of the local geometry around the keypoint. 

\subsection{Rotation-Equivariant Descriptor}
\label{subsec:rotation_equivariant}

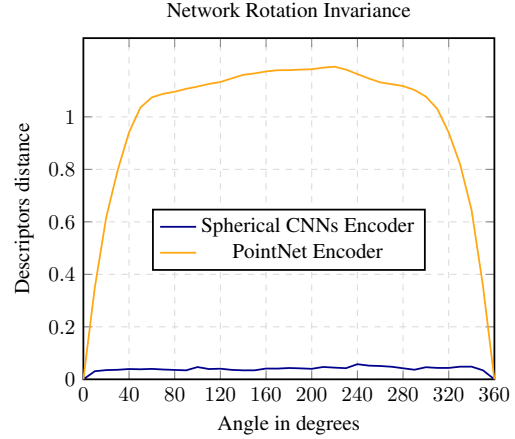
\begin{figure}[t]
	\centering
	\resizebox{0.4\textwidth}{!}{
		\begin{tikzpicture}
\begin{axis}[
legend style={at={(0.5,0.5)},anchor=north},
grid = major,
grid style = {dashed, gray!30},
title={Network Rotation Invariance},
x tick label style={/pgf/number format/fixed},
xmin = 0,
xmax = 360,  
ymin = 0,   
ymax = 1.3, 
ytick={0.0,0.2,...,1.0},
xtick={0,40,...,360},
axis background/.style = {fill=white},
line width=0.85pt,
ylabel = {Descriptors distance},
xlabel = {Angle in degrees}]
\addplot[ours] table {./results/network_rotation_ours.dat};
\addplot[atlasnet] table {./results/network_rotation_point_net.dat};
\addlegendentry{Spherical CNNs Encoder}
\addlegendentry{PointNet Encoder}
\end{axis}
\end{tikzpicture}
	}
	\caption{Comparison between PointNet and Spherical CNN used as encoders in our framework. \label{fig:equivariance}}
\end{figure} 

The main novelty of our approach is the use of Spherical CNNs as encoder to learn an equivariant bottleneck layer.

Learning an equivariant bottleneck removes the requirement to have invariant representations as input to the network at training time as the only way to achieve rotation invariance, the standard approach in existing proposals \cite{deng2018ppffoldnet, khoury2017learning}. In our framework, instead, we can delay the choice on how to canonically orient the descriptor at test time, which brings in two important benefits. On one hand, we do not have to choose a specific way to orient the input, e.g. a specific LRF, at training time, which means that we can train the network to learn the descriptor from less pre-processed input data than existing proposals, moving a step closer toward end-to-end descriptor learning.
On the other hand, not using a LRF at training time frees our method from unavoidable errors of the LRF itself, which in turn inject noise in the training process. We expect both benefits to concur to increase the effectiveness of the learned descriptor.

Moreover, from a practical point of view, being able to train our descriptor without tying it to a specific LRF enables us to choose the best way to define a canonical representation at test time without training the network from scratch. Finally, it also opens up the possibility to use different LRFs for different test data, although we have not explored this property in the experimental results reported in this paper.  

Please note that a truly rotation-equivariant CNN like Spherical CNNs is mandatory in our framework, as anticipated in the introduction. Indeed,  only a descriptor that lives in $\SO{3}$ can be rotated after having been computed, i.e. only the output of a Spherical CNN to date. All the other standard representations, e.g. the output of a Multi Layer Perceptron (MLP) as used in PointNet, cannot be rotated after having been computed. Therefore, if we want to use them in our framework where the input is not canonically oriented for the reasons discussed above, we can only hope the network learns to obtain directly a rotation-invariant descriptor by observing rotated versions of the same neighborhood during training without explicit supervision, which is however a harder task in our setup than learning an equivariant descriptor.

We have validated how harder this is experimentally, by using a standard PointNet encoder  instead of the spherical one to learn an invariant descriptor. 
Results of the comparison are shown in Figure \ref{fig:equivariance}. Please note that equivariance is a theoretical property of a Spherical CNN, regardless of whether it is trained or not. Indeed, in the results in Figure \ref{fig:equivariance}, the Spherical encoder has {\bf not} been trained, while the PointNet encoder has been trained on the 3DMatch Benchmark presented in Section \ref{sec:exp_setup}. Given a neighborhood, we rotate it around a random axis by a growing angle, whose value is reported along the horizontal axis of the chart. For every rotation, we pass the rotated neighborhood through a Spherical CNN encoder and a PointNet encoder. The output of the Spherical CNN is then rotated by the inverse of the applied rotation (simulating the availability of a perfect LRF) and the distance between the descriptor obtained from the rotated neighborhood and the descriptor obtained from the un-rotated neighborhood is plotted. We can clearly see that PointNet fails to learn an invariant descriptor in our setup, while the equivariant representation provided by a Spherical CNN can achieve almost perfect invariance when properly rotated. 

\begin{figure}[h]
	\centering
	\begin{tabular}{ccc}

		\includegraphics[width=.12\textwidth]{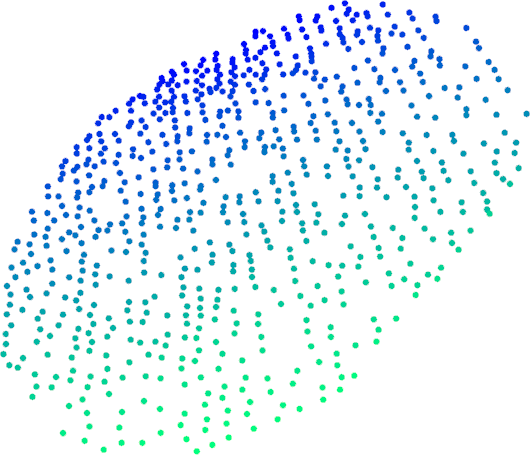}&
		\includegraphics[width=.12\textwidth]{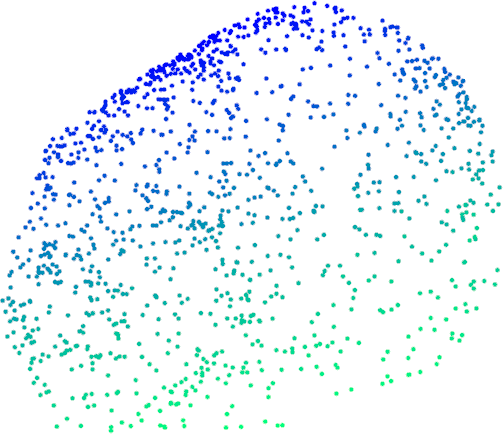}&
		\includegraphics[width=.12\textwidth]{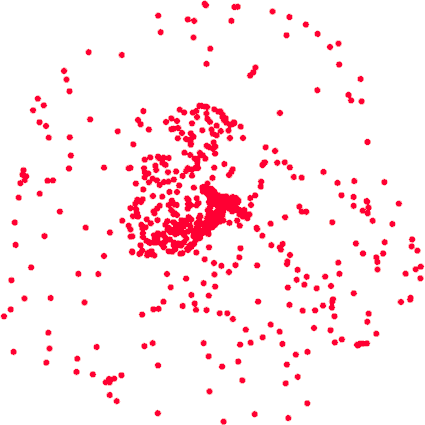}\\
		
		\includegraphics[width=.12\textwidth]{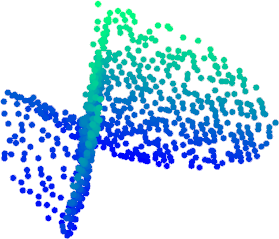}&
		\includegraphics[width=.12\textwidth]{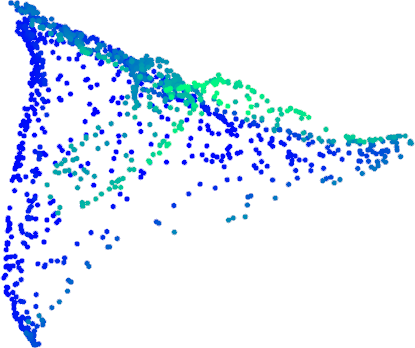}&
		\includegraphics[width=.12\textwidth]{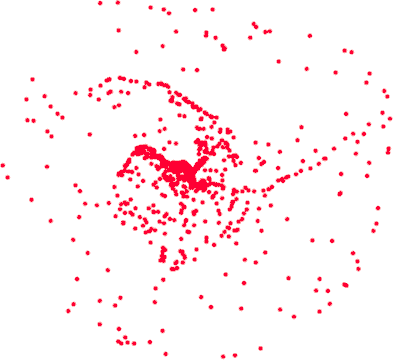}\\
		
		\includegraphics[width=.12\textwidth]{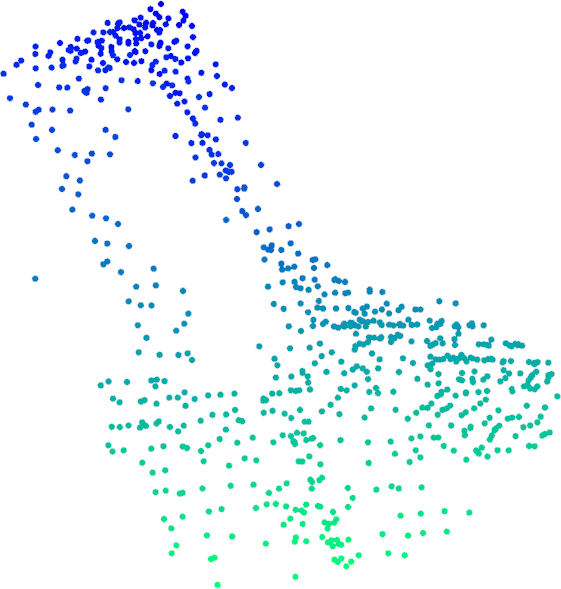}&
		\includegraphics[width=.12\textwidth]{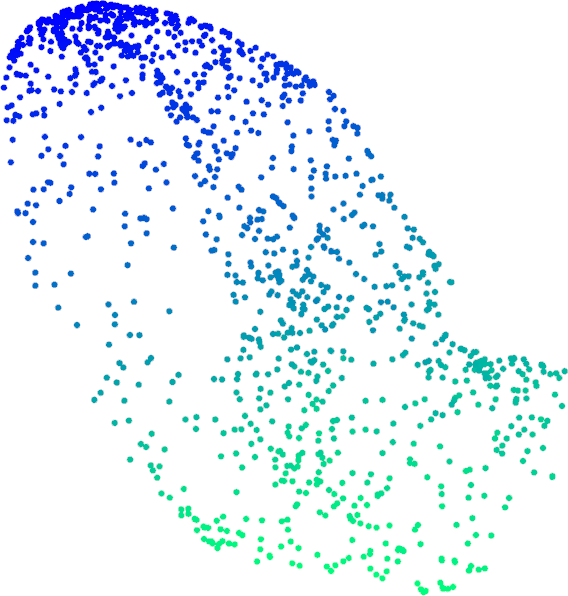}&
		\includegraphics[width=.12
		\textwidth]{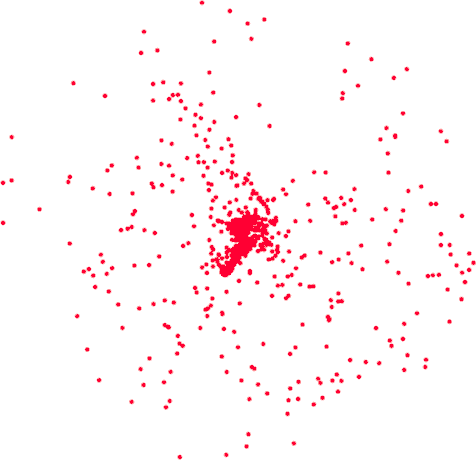}\\
				
		\small Input &
		\small Equivariant  &
		\small Invariant  \\
	\end{tabular}
	\caption{Comparison between the reconstructions obtained when using the Spherical CNN encoder to learn an equivariant versus an invariant bottleneck. Results after 10K training iterations.}
	\label{fig:training}
\end{figure}

Moreover, even if PointNet were able to learn a perfectly invariant bottleneck, we have found experimentally that this would result in low quality reconstructions. The reason is that it is not possible for frameworks like FoldingNet/AtlasNet to converge to sensible reconstructions if the learned bottleneck does not contain any pose information, i.e. it is almost perfectly invariant. This is shown in Figure \ref{fig:training}. where we compare the quality of the reconstructions produced by our framework when using an equivariant bottleneck layer versus an invariant one. The invariant one in this case is obtained by removing the last $\SO{3}$ correlation layer from our encoder, which produces the equivariant descriptor in our architecture, and adding a max pooling layer selecting the maximum of each one of the now top-level 40 feature maps, followed by a fully connected layer to expand the codeword dimensionality to 512. As shown in the figure, if the encoder produces an invariant descriptor, the decoder doesn't have enough information to know in which pose it should reconstruct the input so as to minimize the loss. The best it can do is to produce reconstructions trying to account for all possible rotations of the input, e.g. the atom-like structures depicted in the last column, almost ignoring the invariant bottleneck layer.

\subsection{Invariant Feature Descriptor}

\label{subsec:equivariant_to_invariant}

To obtain an invariant descriptor at test time, which can be matched across poses, we have to compute a canonical orientation for the equivariant descriptor. We have investigated two ways of doing it.

The first is the most intellectually satisfying, and leverages again the peculiar properties of Spherical CNNs. Indeed, every bin of a feature map in a Spherical CNN represents an element of $\SO{3}$, \ie a potential LRF. This has been already exploited to align full shapes in \cite{esteves2018learning}, by finding the $\argmax$ of the correlation between two feature maps. Note that we cannot use the same approach in the context of invariant descriptor matching, as this would require a costly computation to compute the distance between every pair of source and target descriptors.

However, because of the equivariance property, we can recover an aligning pose by processing the two descriptors separately. Let $[\psi \ast h](R)$ be the descriptor, \ie a feature map, obtained when processing the input signal $f$, and let $[\psi \ast m](R)$ the one obtained when we process a rotated version of $f$, $g(x) = [L_Q f] (x) = f(Q^{-1}x)$. Due to equivariance, the same rotation exists between inner feature maps $h$ and $m$, \ie $m(R) = [L_Q h] (R) = h(Q^{-1}R)$, and recursively between descriptors, \ie
\begin{align}
\label{eqn:equivariance}
[\psi \ast m ](R_m) & =  [\psi \ast [ L_Q h] ](R_m)  \nonumber\\
& = \langle L_{R_m} \psi, L_Q h \rangle \nonumber \\
& =  \langle L_{Q^{-1} R_m} \psi, h \rangle \nonumber \\
& = [\psi \ast h ](Q^{-1} R_m) \coloneqq [\psi \ast h ](R_h) 
\end{align}

\begin{figure}
	\centering
	\includegraphics[width=0.35\textwidth]{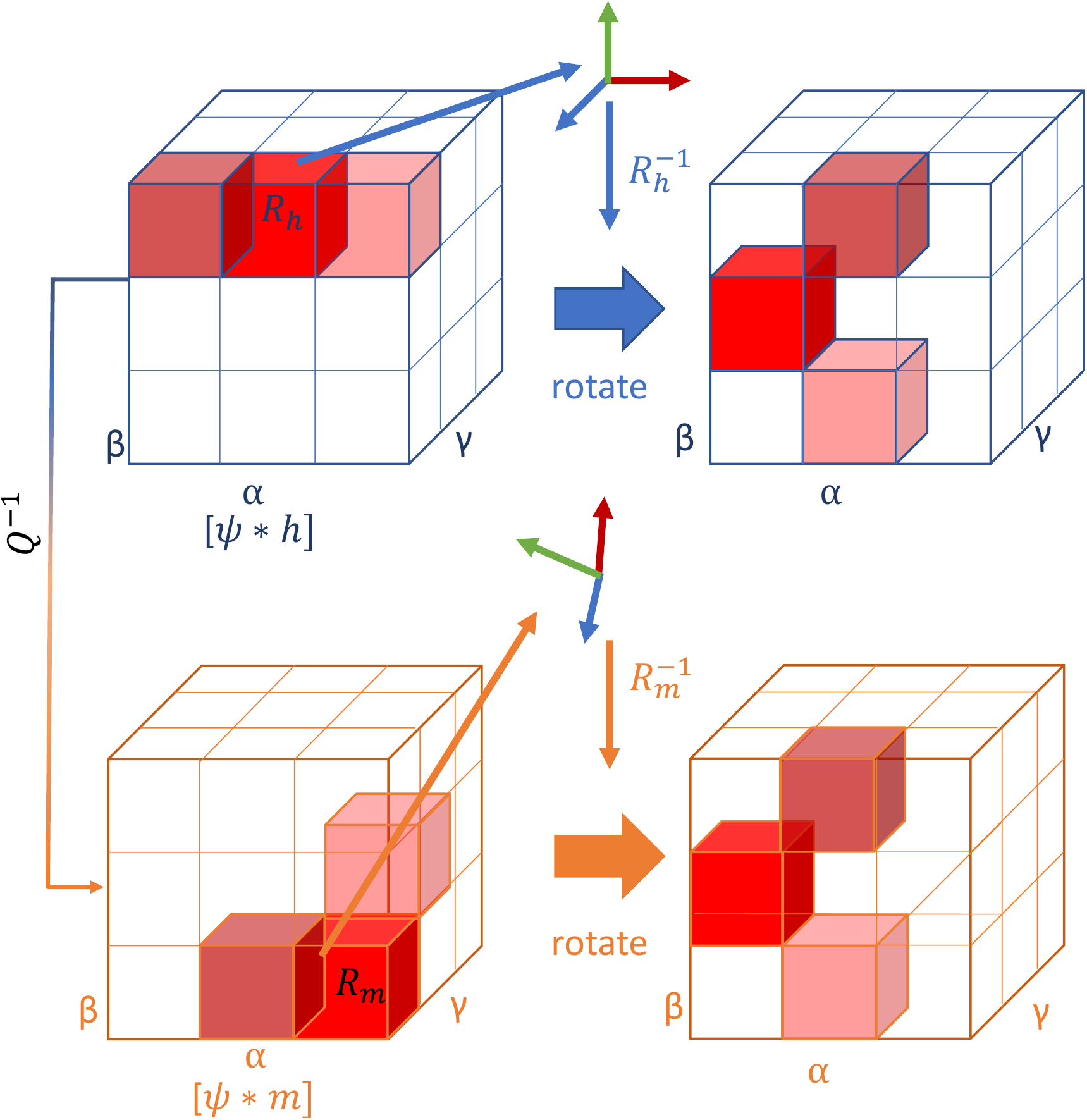}
	\caption{Self-orienting property of the learned equivariant descriptor. Every bin of our bottleneck layer corresponds to three Euler angles which define a rotation. If the descriptor is computed starting from a rotated input (second row), the values shifts in the feature maps. By finding two corresponding bins in the two descriptors and rotating them by the inverse of the corresponding rotations, the descriptors can be aligned, i.e. become pose invariant. \label{fig:lrf}}
\end{figure}

In other words, chosen an entry in a descriptor $[\psi \ast h ]$ obtained when processing $f$, e.g. $R_h$, if, when the input is rotated by $Q$, we are able to find the same entry independently in the rotated descriptor $[\psi \ast m]$, we will find it at rotation $R_m = Q R_h$.  
Therefore, given the two descriptors, we can align them to a common pose by applying the inverse of such rotations
\begin{align}
[L_{R_m^{-1}}[\psi \ast m ]] (R) & = [\psi \ast m ] (R_m R) \nonumber \\
&=  [\psi \ast [ L_Q h ] ] (R_m R) \nonumber \\
& = [\psi \ast h ] (Q^{-1} R_m R) \\
[L_{R_h^{-1}}[\psi \ast h ]] (R)&  = [\psi \ast h ] (R_h R) \nonumber \\ 
& = [\psi \ast h ] (Q^{-1} R_m R)
\end{align} 
as shown by last terms of the transformations being equal (and graphically in Figure \ref{fig:lrf}). 
Please note that all transformations are applied to the descriptor (which is a feature map) obtained from the unrotated input and not to the input itself, i.e. we can rotate the descriptor computed from an unoriented input to achieve rotation invariance. The dimensionality of ours descriptor does not change under rotations, as it is rotated by remodulating the spherical harmonics functions resulting from its Fourier transform. A through treatment of this topic can be found in \cite{risbo1996fourier}.

The problem of defining a repeatable LRF then translates into that of finding the same bin under rotations given a feature map. A simple choice could be the maximum of the feature map. Under perfect equivariance, the maximum would provide a repeatable anchor point across rotations, and therefore a repeatable rotation to obtain invariant descriptors. However, the network is not perfectly equivariant, due to numerical approximations and the use of non linearities (ReLUs) between layers, and also the feature map of the same keypoint seen in two different views changes due to other nuisances (occlusions, clutter, sampling). We have verified experimentally that the maximum of a feature map alone is not robust enough to define a repeatable LRF.

We have investigated several strategies to identify the same location of the feature map under rotations. The one that has given the best results so far starts by analyzing only the $k$ bins corresponding to the top $k$ values of the feature map, including the maximum. We then compute the density of top values in a $3 \times 3 \times 3$ neighborhood of every such bin. The bin with the maximum density is used to compute the required rotation. In the case of ties, we select the neighborhood with the largest value within it. Once we have selected a neighborhood, we average all the bins corresponding to the top values within it to get the final rotation. By means of the proposed algorithm, we have been able to define a \textit{self-orienting} descriptor, an original trait of our proposal.
 
As our tests indicate the repeatability of the above defined LRF to be far from the optimal performance attainable with the equivariant descriptor, we have also assessed its performance when we make it invariant at test time by computing the canonicalizing rotation with the help of an external local reference frame extracted from the input cloud. We stress here that, although we compute an LRF on the input data, we again rotate the computed descriptor and not the input data. Moreover, even in this case, we perform LRF extraction only at test-time, as discussed above, so the chosen LRF algorithm does not affect the quality of the training data.

\subsection{Decoder and Loss}
\label{subsec:decoder}
Differently from \cite{deng2018ppfnet}, our goal is to reconstruct the whole set of points representing the local neighborhood of a given feature point $\pp$. Inspired by \cite{groueix2018papier} and \cite{yang2018foldingnet}, our decoder will try to deform points in $ \mathbb{R}^2$ to surface points in $ \mathbb{R}^3$ according to the learned descriptor. Given a feature representation $\dd$ for a 3D surface, let $\mathcal{A}$ be a set of points sampled in the unit square $[0,1]^2$, the descriptor $\dd$ is concatenated with the sampled point coordinates $(a_x, a_y) \in \mathcal{A}$ and then forwarded through a stack of MLP layers as shown in Figure \ref{fig:architecture}. We then minimize the Chamfer loss between the set of generated 3D points and the input points. 

In particular, let $\source$ be the set of 3D input points belonging to the neighbohood of $\pp$ and $\reconstructed$ the set of points reconstructed by the decoder. During training, we minimize the following loss 
\begin{equation}
\label{eqn:loss}
\begin{split}
\loss(\source,\reconstructed)_{\parameters} = \frac{1}{|\source|}\sum\limits_{\mathbf{x}\in \source}\min_{{\mathbf{x}^\star}\in {S}^\star}\|\mathbf{x}-{\mathbf{x}^\star}\|_2 +
\\ 
\frac{1}{|\reconstructed|}\sum\limits_{{\mathbf{x}^\star}\in \reconstructed}\min_{\mathbf{x}\in \source}\|{\mathbf{x}^\star} - \mathbf{x}\|_2.
\end{split}
\end{equation}
The term $\min_{{\mathbf{x}^\star}\in {S}^\star}\|\mathbf{x}-{\mathbf{x}^\star}\|_2$ enforces that any 3D point $\mathbf{x}$ in the original point cloud has a matching 3D point ${\mathbf{x}^\star}$ in the reconstructed point cloud, and the term $\min_{\mathbf{x}\in S}\|{\mathbf{x}^\star} - \mathbf{x}\|_2$ enforces the matching viceversa. The overall loss is the sum of the two terms to enforce that the distance from $\source$ to $\reconstructed$ and the distance viceversa have to be small simultaneously. 

\subsection{Network and training parameters}
To learn our descriptor, we use one $S^2$ convolution layers and three $\SO{3}$ convolution layers with constant number of channels, 40, while the bandwidths is set to 24 for the first three layer and 4 for the last one, which results in a descriptor with 512 entries. 
The architecture of our decoder is made of 4 fully-connected layers, with ReLU non-linearities on the first three layers and tanh on the final output layer.
The network is trained with mini-bacthes of size 32 by using ADAM \cite{kingma2014adam}. The starting learning rate is set to 0.001 and is decayed every 4000 iterations. We train the network for 14 epochs.

\section{Experimental Results}
\label{sec:results}

\begin{table*}[t!]
	\centering
	\caption{Results on the 3DMatch benchmark. Test data are from SUN3D \cite{xiao2013sun3d}, except for \textit{Red Kitchen} data which is from 7-scenes \cite{shotton2013scene}. Best result on each row is in bold.}
	\resizebox{\textwidth}{!}{
		\begin{tabular}{lcccccccccc}
			\hline
			& FPFH \cite{rusu2009fast} & Spin Image~\cite{johnson1999using} & SHOT~\cite{salti2014shot} & USC \cite{tombari2010unique} & 3DMatch~\cite{zeng20173dmatch} & CGF~\cite{khoury2017learning} & PPFNet~\cite{deng2018ppfnet} & PPFFoldNet \cite{deng2018ppffoldnet} & Ours SO & Ours LRF \\
			\hline
			Kitchen & 0.7391 & 0.6561 & 0.8893 & 0.9308 & 0.5810 & 0.4605 & 0.8972 & 	0.7866 	& 0.8854 & \bf{0.9763} \\
			Home 1  & 0.7885 & 0.7564 & 0.8974 & 0.9103 & 0.7244 & 0.6154 & 0.5577 & 	0.7628 	& 0.9487 & \bf{0.9615} \\
			Home 2  & 0.6442 & 0.6731 & 0.8221 & 0.7788 & 0.6154 & 0.5625 & 0.5913 & 	0.6154 	& 0.8654 & \bf{0.8942} \\
			Hotel 1 & 0.8142 & 0.6770 & 0.9336 & 0.9204 & 0.5442 & 0.4469 & 0.5796 & 	0.6814 	& 0.9204 & \bf{0.9823} \\
			Hotel 2 & 0.7115 & 0.6346 & 0.8750 & 0.8462 & 0.4808 & 0.3846 & 0.5796 & 	0.7115 	& 0.8462 & \bf{0.9519} \\
			Hotel 3 & 0.8889 & 0.7407 & 0.8889 & 0.8889 & 0.6111 & 0.5926 & 0.6111 & 	0.9444	& 0.9630 & \bf{0.9815} \\
			Study   & 0.7432 & 0.4692 & 0.8630 & 0.8664 & 0.5171 & 0.4075 & 0.5342 & 	0.6199 	& 0.8870 & \bf{0.9178} \\
			MIT Lab & 0.7013 & 0.4545 & 0.8312 & 0.8052 & 0.5065 & 0.3506 & 0.6364 & 	0.6234 	& 0.8182 & \bf{0.8701} \\
			\hline                                                                                   
			Average & 0.7539 & 0.6327 & 0.8751 & 0.8684 & 0.5726 & 0.4776 & 0.6231 & 	0.7182 	& 0.8918 & \bf{0.9420} \\
			\hline
		\end{tabular}%
		\label{tab:3dmatchbenchmark}%
	}
\end{table*}%

\begin{table*}[t!]
	\centering
\caption{Results on the rotated 3DMatch benchmark. Test data are from SUN3D \cite{xiao2013sun3d}, except for \textit{Red Kitchen} data which is from 7-scenes \cite{shotton2013scene}. Best result on each row is in bold.}
\resizebox{\textwidth}{!}{
\begin{tabular}{lccccccccccc}
	\hline
	& FPFH \cite{rusu2009fast} & Spin Image~\cite{johnson1999using} & SHOT~\cite{salti2014shot} & USC \cite{tombari2010unique} & 3DMatch~\cite{zeng20173dmatch} & CGF~\cite{khoury2017learning} & PPFNet~\cite{deng2018ppfnet} & PPFFoldNet \cite{deng2018ppffoldnet} & Ours SO & Ours LRF \\
	\hline
	Kitchen & 0.7451 & 0.6502 & 0.8794 & 0.9170 & 0.004  & 0.4466 & 0.002  & 0.7885 	& 0.8893 & \bf{0.9783} 	\\
	Home 1  & 0.7949 & 0.7628 & 0.8910 & 0.9103 & 0.0128 & 0.6667 & 0.0000 & 0.7821 	& 0.9423 & \bf{0.9679} 	\\
	Home 2  & 0.6587 & 0.6635 & 0.8317 & 0.7548 & 0.0337 & 0.5288 & 0.0144 & 0.6442 	& 0.8413 & \bf{0.8894} 	\\
	Hotel 1 & 0.8142 & 0.6903 & 0.9425 & 0.9292 & 0.0044 & 0.4425 & 0.0044 & 0.6770 	& 0.9204 & \bf{0.9779} 	\\
	Hotel 2 & 0.7212 & 0.6635 & 0.8654 & 0.8558 & 0.0000 & 0.4423 & 0.0000 & 0.6923 	& 0.8558 & \bf{0.9615} 	\\
	Hotel 3 & 0.9259 & 0.7222 & 0.9074 & 0.9074 & 0.0096 & 0.6269 & 0.0000 & 0.9630		& 0.9074 & \bf{0.9815} 	\\
	Study   & 0.7260 & 0.4692 & 0.8493 & 0.8836 & 0.0000 & 0.4178 & 0.0000 & 0.6267 	& 0.8733 & \bf{0.9110} 	\\
	MIT Lab & 0.7532 & 0.4935 & 0.8312 & \bf{0.8571} & 0.0026 & 0.4156 & 0.0000 & 0.6753 	& 0.7922 & 0.8442 	\\
	\hline                                                                                            
	Average & 0.7674 & 0.6394 & 0.8747 & 0.8769 & 0.0113 & 0.4776 & 0.0026 & 0.7311 	& 0.8778 & \bf{0.9387}  \\
	\hline
\end{tabular}%
	\label{tab:rotated3dmatchbenchmark}%
}
\end{table*}%

\subsection{Experimental setup}
\label{sec:exp_setup}
To test our proposal, we use the standard benchmark for the evaluation of learned 3D descriptors, the 3DMatch benchmark \cite{zeng20173dmatch}. This benchmark addresses registration of unordered 3D views
 and the dataset has been put together by merging a large part of the publicly available datasets such as Analysis-by-Synthesis \cite{valentin2016learning}, 7-Scenes \cite{shotton2013scene}, SUN3D \cite{xiao2013sun3d}, RGB-D Scenes v.2 \cite{lai2014unsupervised} and Halberand Funkhouser \cite{halber2017fine}. It contains 62 scenes in total, and, following \cite{deng2018ppffoldnet}, we use 54 for training and validation, while 8 scenes are used only at test time to run comparisons. The dataset already provides so-called fragments, \ie the point clouds resulting from the fusion of 50 consecutive depth frames, for the test scenes, and we obtained the training fragments generated by the same methodology as the authors of \cite{deng2018ppffoldnet}.
We also procure the rotated version of the 3D Match benchmark, generated by the same authors by rotating all the fragments in the 3DMatch benchmark with randomly sampled axes and angles over the whole rotation space. 

We use the same setup proposed in \cite{deng2018ppffoldnet}: we downsample the fused fragments with a voxel grid filter of size 2 cm and compute surface normals using \cite{hoppe1992surface} in a 17-point neighborhood; we consider a radius of 30 cm to define the neighborhood of a keypoint.

\subsection{Evaluation methodology}
As for metrics, following the evaluation methodology proposed by \cite{deng2018ppffoldnet}, we consider the \textit{recall} of pairs of fragments correctly registered among those with at least $30\%$ overlap. A pair of fragments is considered correctly registered if the number of correctly matched keypoints is greater than the inlier ratio threshold $\tau_2$, set to $5 \%$ of the extracted keypoints. Two keypoints are correctly matched if their $l_2$ distance is below a threshold $\tau_1 = 10$ cm. For each fragment, the descriptors are computed on 5000 uniformly sampled points, provided with the benchmark \cite{zeng20173dmatch}. 
For handcrafted descriptors we used the implementation in PCL \cite{aldoma2012tutorial}, while for learned descriptors results were taken from \cite{deng2018ppffoldnet}.

\subsection{Quantitative results}

Results of the tests on the 3D Match benchmark in terms of recall are reported in Table \ref{tab:3dmatchbenchmark}.
With Ours SO we refer to the self orienting descriptor introduced in section \ref{subsec:equivariant_to_invariant}, while with Ours LRF we refer to the descriptor oriented with an external local reference frame. In particular for this experiments we have used the LRF algorithm proposed in \cite{petrelli2012repeatable}, which we will denote as FLARE according to the acronym used in its PCL implementation \cite{aldoma2012tutorial}.

\begin{figure}[t]
	\centering
	\resizebox{0.45\textwidth}{!}{
	\begin{tikzpicture}
\begin{axis}[
grid = major,
grid style = {dashed, gray!30},
title={3D MatchBenchmark},
x tick label style={/pgf/number format/fixed},
legend pos=outer north east,
legend cell align=left,
xmin = 0.0,
xmax = 0.20,  
ymin = 0,   
ymax = 1.0, 
ytick={0.0,0.2,...,1.0},
xtick={0.0,0.04,0.08,0.12,0.16,0.20},
axis background/.style = {fill=white},
line width=0.85pt,
ylabel = {Recall },
xlabel = {Inlier ratio threshold}]
\addplot[ours_flare] table {./results/3DMatch/ours_flare.dat};
\addplot[ours] table {./results/3DMatch/ours_so.dat};
\addplot[fpfh] table {./results/3DMatch/fpfh.dat};
\addplot[shot] table {./results/3DMatch/shot.dat};
\addplot[spinimages] table {./results/3DMatch/spin_images.dat};
\addplot[usc] table {./results/3DMatch/usc.dat};
\addplot[3dmatch] table {./results/3DMatch/3dmatch.dat};
\addplot[cgf] table {./results/3DMatch/cgf.dat};
\addplot[ppfnet] table {./results/3DMatch/ppf.dat};
\addplot[ppffoldnet] table {./results/3DMatch/ppf_foldnet.dat};
\addlegendentry{Ours LRF}
\addlegendentry{Ours SO}
\addlegendentry{FPFH}
\addlegendentry{SHOT}
\addlegendentry{Spin Images}
\addlegendentry{USC}
\addlegendentry{3D Match}
\addlegendentry{CGF}
\addlegendentry{PPFNet}
\addlegendentry{PPF-FoldNet}
\end{axis}
\end{tikzpicture}}
	\caption{Results under varying inlier ratio threshold $\tau_2$. \label{fig:3dmatch}}
\end{figure}
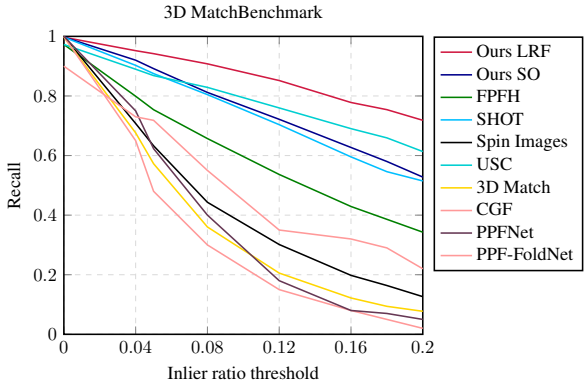

The first outcome of our experiments is that the use of an external LRF outperforms the self-orienting variant of our algorithm. Note that the two columns describe exactly the same equivariant descriptor under two different ways to compute a canonical orientation. Hence, the highest one is indicative of the quality of the learned descriptor itself. Although the performance of the self-orienting variant of our method is inferior to our descriptor oriented by an external LRF, it is remarkable that it delivers the second best recall on the dataset, \ie it would provide state-of-the-art performance if we were not to orient our equivariant descriptor also with an external LRF. Our self-orienting variant is closely followed by SHOT and USC, \ie two handcrafted descriptors, while the other tested methods deliver significantly lower recalls. The best learned approach is PPFFoldNet. The better performance of SHOT and USC with respect to PPFFoldNet offers support to the inspiring ideas behind this work: deep learning alone, if constrained to learn from highly engineered representations, cannot be a guarantee of superior performance.
It is also interesting to analyze these results in light of our main claim: to learn an equivariant descriptor and then orient it to achieve invariance instead of learning directly an invariant one boosts its quality. If we compare the performance of our method when oriented with both tested variants against methods learning from invariant representations, like PPFFoldNet and CGF, we can interpret the large gap in performance (0.23 and 0.47 points of recall from the external LRF variant, respectively), as a validation of the drawbacks of existing learned descriptors discussed in the introduction.
In Figure \ref{fig:3dmatch}, we report results when varying the threshold $\tau_2$ on the percentage of correct matches to consider a pair as correctly registered, as done in \cite{deng2018ppffoldnet}. Our proposal oriented with an external LRF outperforms the others for all thresholds, and our self-orienting variant attains again recall values similar to SHOT, and slightly inferior to USC for the largest thresholds. 

Finally, results of the tests on the rotated 3D Match benchmark are reported in Table \ref{tab:rotated3dmatchbenchmark}. The dataset was proposed in \cite{deng2018ppffoldnet} to test robustness against large rotations, not present in the original benchmark. As expected, all rotation-invariant methods obtain performance similar to the results reported in Table \ref{tab:3dmatchbenchmark}, and our equivariant descriptor oriented with the external LRF still delivers by far the best performance.

\section{Conclusions}
\label{sec:conclusion}
In this study, we have shown how the problem of learning an effective descriptor can be separated into the orthogonal problems of learning a robust equivariant representation and defining a good canonical orientation to make it invariant at test time. 
Our proposal to learn an equivariant representation in an unsupervised way leverages as encoder the recently proposed Spherical CNNs and turns out highly effective in tackling the first  problem. When coupled with a robust algorithm to compute a local reference frame from the input cloud, it significantly advances the state of the art on a challenging benchmark. 

We have also shown how the very same framework could be used to define a canonical orientation by exploiting the peculiar nature of the feature maps computed by the Spherical CNNs. Although this approach delivers performance on par with the state of the art, it is so far inferior to the use of an external LRF. Yet, we believe the elegance and potential implications of this technique were valid reasons to also communicate it and call for further studies along this line of research, with the aim of defining an end-to-end learned solution to the problem of invariant 3D description.

\section{Acknowledgments}
We gratefully acknowledge NVIDIA Corporation with the donation of the Titan V GPU used in this work.
{\small
\bibliographystyle{ieee_fullname}
\bibliography{egbib}
}

\end{document}